# Enhanced Frame and Event-Based Simulator and Event-Based Video Interpolation Network

Adam Radomski[1], Andreas Georgiou[1], Thomas Debrunner[1], Chenghan Li[1], Luca Longinotti[1], Minwon Seo[2], Moosung Kwak[2], Chang-Woo Shin[2], Paul K. J. Park[2], Hyunsurk Eric Ryu[3], Kynan Eng[1]

[1]*iniVation AG*
Zurich, Switzerland
info@inivation.com

[2]*Samsung Electronics*
1-1 Samsungjeonja-ro, Hwaseong-si,
Gyeonggi-do 445-330, South Korea
k.j.park@samsung.com

[3]*Seoul National University*
Seoul, South Korea
eric.ryu@snu.ac.kr

*Abstract*—Fast neuromorphic event-based vision sensors (Dynamic Vision Sensor, DVS) can be combined with slower conventional frame-based sensors to enable higher-quality inter-frame interpolation than traditional methods relying on fixed motion approximations using e.g. optical flow. In this work we present a new, advanced event simulator that can produce realistic scenes recorded by a camera rig with an arbitrary number of sensors located at fixed offsets. It includes a new configurable frame-based image sensor model with realistic image quality reduction effects, and an extended DVS model with more accurate characteristics. We use our simulator to train a novel reconstruction model designed for end-to-end reconstruction of high-fps video. Unlike previously published methods, our method does not require the frame and DVS cameras to have the same optics, positions, or camera resolutions. It is also not limited to objects a fixed distance from the sensor. We show that data generated by our simulator can be used to train our new model, leading to reconstructed images on public datasets of equivalent or better quality than the state of the art. We also show our sensor generalizing to data recorded by real sensors.

## I. INTRODUCTION

Since the invention of neuromorphic event-based vision sensors such as the Dynamic Vision Sensor (DVS) [1] and the variants with global shutter frames [2], many algorithms and use cases for such sensors have been developed [3]. A key issue in the development of algorithms for such sensors is the inherently lossy nature of the output data. In most conventional sensors the design goal is to minimize this information reduction, while in event-based sensors the information reduction is exploited to achieve gains in speed and energy efficiency. Because of the pixel-level noise inherent in the binary event generation mechanism, small changes in noise and signal can lead to large changes in algorithm performance. Therefore, it is important to be able to vary sensor and input parameters in a systematic way to test the robustness of any event-based algorithm. Event-based simulators can be an effective, efficient way to achieve this goal.

### A. Event Sensor Simulators

A number of DVS simulators have been created over the years, with differing approaches to noise modelling and input scene generation. The **ESIM** simulator [4] is constructed around a fully synthetic 3D environments. Its noise model includes a Gaussian event threshold mismatch. It also includes a frame simulator including a motion blur model and a simulated camera-mounted inertial measurement unit (IMU). Because it is fully synthetic, photorealistic scenes require significant effort to create, and real visual data cannot be easily mapped directly into the simulator.

The **Video to Events** simulator [5] avoids the need to render custom event-based datasets by using existing videos. To provide the required time resolution, input video is temporally upsampled before being fed to ESIM. The quality of the events is thus dependent on the quality of the video upsampling method.

The **v2e** simulator [6] also converts videos into events, but has more features in its noise model: pixel-level Gaussian event threshold mismatch, finite intensity-dependent bandwidth, and intensity-dependent noise. In another development, it has been shown that when a model of a deformable object is available, a differentiable renderer can be used to combine predicted and observed events to predict object pose [7] – however this work did not consider noise in its model.

In our work, we introduce a new frame-and-event simulator with several new features to emulate realistic DVS and frame-based camera characteristics. Its DVS features include pixel-level event threshold mismatch, intensity-dependent bandwidth and noise, refractory period noise, and faulty pixel simulation. It also includes frame output effects such as pixel-level noise and rolling shutter.

### B. Image Reconstruction Using Event Sensors

Because event-based sensors convert signal resolution into the temporal domain, it is theoretically possible to recover the signal. While such a recovery process is necessarily limited due to effects such as event generation refractory period and limited time resolution, it is possible in the general case to recover grey-level scene information. Such a recovery process allows a wider range of existing algorithms to be used on the event data, and also allows for easier human interpretation of the scene.

Although there are many simple event integration-based methods for reconstructing images, in this paper we consider only more advanced methods using neural networks. To evaluate the quality of image reconstruction we used both the peak signal-to-noise ratio (PSNR) and the structural similarity index measure (SSIM) [8].

**Events-to-Video (E2VID)** [9] was one of the first published video reconstruction networks which produced reasonable quality output. The network uses a single recurrent UNet which memorizes the scene appearance from the previous reconstructions and updates it with each newly arrived event bin. The original network architecture does not take any CIS (CMOS Image Sensor) frames as an input.

In our testing of E2VID, we trained the network on data generated by our event simulator. For the loss function we used a combination of L1 and perceptual loss. The network was unrolled for the training and the loss function was calculated for all but the first reconstructed frames. We achieved similar results to those in the original paper. The reconstructed frames were of generally poor quality, reaching a SSIM of only 0.56.

**Fast Image Reconstruction Network** (FireNet) [10] is an improvement of the E2VID network focused on reducing the number of parameters (38k vs 10M) and making the reconstruction of a single frame faster (10ms vs 30ms). In our testing, we implemented, trained and tested the network on data generated by our event simulator. For training we used a loss function similar to the one described in the original paper which is a combination of L1 loss, perceptual loss and temporal consistency loss. The results were not of very high quality. The SSIM score dropped down to 0.5 and the training was very unstable, and heavily depended on the quality of the training data.

**Time Lens** [16] is a CNN framework which generates videos with high temporal resolution using information from a hybrid camera rig combining an event camera and a RGB camera. The cameras are synchronized in time by the hardware, but the spatial alignment of pixels is done manually. In addition, the resolution of the frames and events is the same, so that the network (ideally) does not have to learn event-to-frame neighborhood relationships.

The current methods have been proven to solve some specific scenarios of image reconstruction, but they still do not work on a generic case and require a lot of manual work to be done on the datasets. In a practical image reconstruction solution, we would expect the disparity issue to be automatically resolved without any additional input from the user with a reasonable computational complexity. Some existing solution datasets also vary the input data frame rate varying from 20 to 60 fps depending on the speed of motion in the scene, which in practice is difficult to predict in advance.

In this paper, we present a new event simulator and image reconstruction framework, target generic event+image reconstruction problems. Our reconstruction network extends the state of the art in the following ways:

- End-to-end image reconstruction solution used in a dual-camera setup with a baseline, using only a stereo calibration as a pre-requirement.

- Input frame rate of the frame camera is fixed to 30fps, which is very practical in real world implementations on hardware.

- Automated alignment between event and frame fields of view using deep learning

- No special constraints on camera motion, no need to display any calibration patterns on the collected videos

- No general restrictions on the relative resolutions of the frame and event sensors

## II. EVENT SIMULATOR

The design goal for the event simulator was to provide a highly adjustable event-based sensor model, combined with a full 3D environment which could provide photorealistic rendering without large effort. Although photorealistic rendering is possible using full 3D models, the creation of suitable content is extremely labor-intensive. It was thus decided to allow users to map high-resolution textures to 3D objects with custom lighting, providing high-fidelity scenes suitable for algorithm development at much lower computational cost and with much lower effort.

### A. Simulator design

A modular overview of the simulator is shown in Fig. 1. The input image datasets should be photorealistic, and can be provided by the user or extracted from publicly available datasets. These photorealistic images are treated as textures of randomly generated 3D objects with various shapes and dimensions. We also made it possible to include any 3D object from external files as part of our scenes for closer representation of the real-world. We also provide an option to configure illumination (including dynamic elements such as sine wave flicker, reflections) of the scene. Objects, light sources and cameras are placed arbitrarily in the virtual volume while moving along predefined or randomly generated trajectories. The view observed by a virtual camera with user-specified characteristics is generated and rendered by an external rendering engine Blender [4]. The rendered images are then processed by our sensor simulation algorithms using the predefined characteristics of each sensor in the camera rig. The components marked in blue are authored as a part of this work and the elements marked in green are used as integrated external tools.

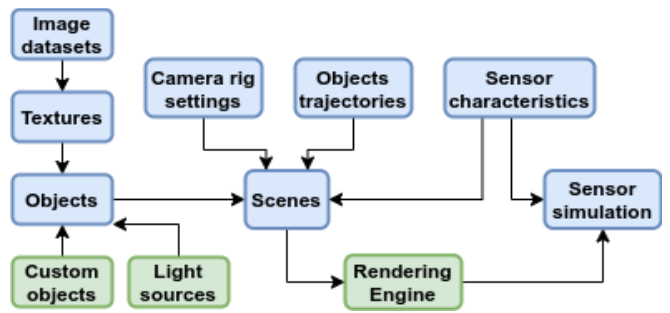

Fig. 1. Event simulator architecture

The simulator can be controlled via GUI (Fig. 2 and Fig. 3) what enables intuitive, rapid building of 3D models with full control over all the parameters. For more convenient usage, we provide 8 types of predefined scenes which can be used to generate any amount of data with just a couple of clicks.

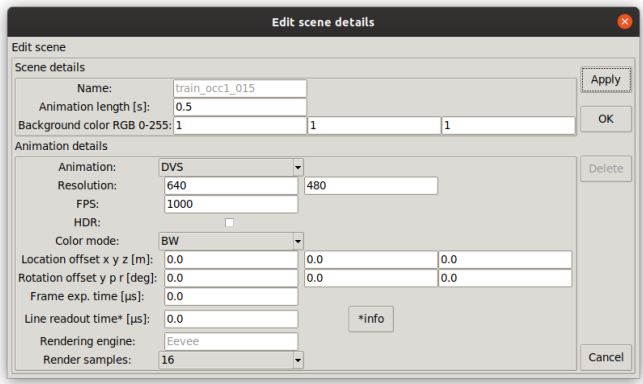

Fig. 2. Event simulator parameter setting GUI example.

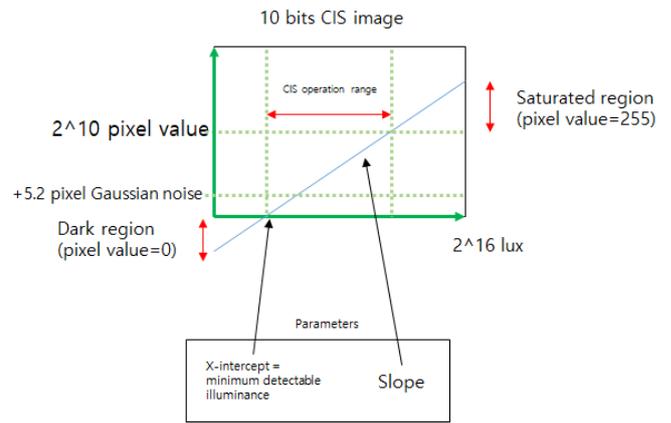

Fig. 4. CIS post-processing model.

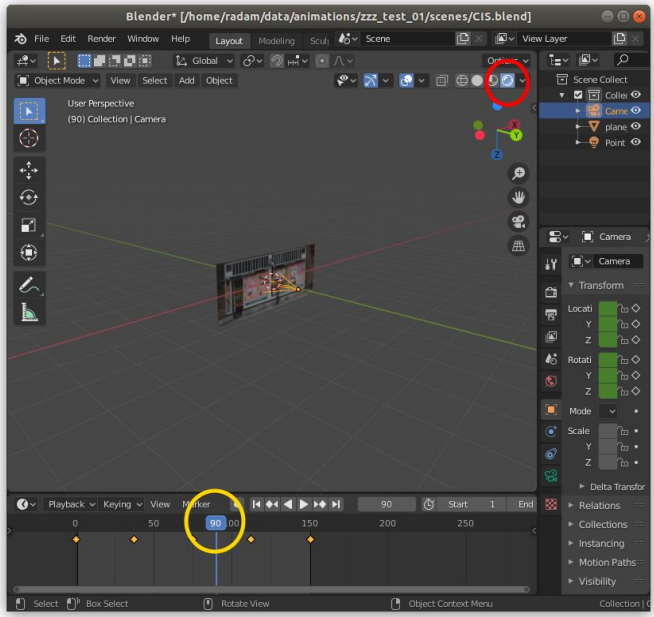

Fig. 3. Event simulator Blender 3D environment view.

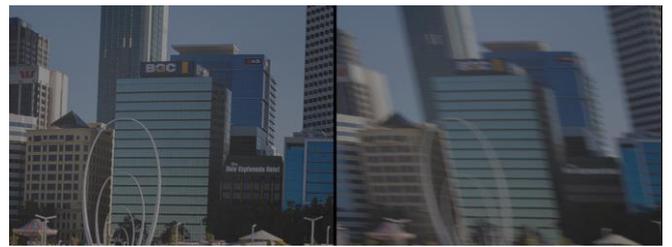

Fig. 5. Example motion blur simulated output incorporating exposure time and line readout time.

The virtual camera rig can model both frame-based CMOS image sensors (CIS) and event-based Dynamic Vision Sensors (DVS). The number of cameras in the rig can be arbitrarily large. Each of the cameras can be shifted relative to the rig frame by a user-specified translation and rotation that can simulate any baseline between camera sensors in case of dual camera hardware. The simulator models the following CIS characteristics (see also Fig. 4 and Fig. 5):

- **Spatial resolution** (pixels, width x height)
- **Time resolution** (frames per second)
- **Motion blur:** Independently controllable exposure time and rolling shutter (line readout time)
- **CIS operation range**: the intensity range in which the CIS operates, incorporating minimum detectable difference and intensity slope
- **Noise:** defined as the number of least significant bits of a 10-bit ADC. A Gaussian noise distribution is assumed.

The sensor simulator models the following DVS characteristics:

- **Spatial resolution** (pixels, width x height)
- **Time resolution** (event-frames per second)
- **Mean event threshold**, for positive and negative events separately, as a floating point number
- **Pixel mismatch**: modeled as time-invariant Gaussian noise with a given standard deviation. The noise distribution is calculated every time the frame to event conversion is executed. In practice, this means that each event sequence generated by the simulator looks as if it was recorded with a unique camera. Specified as a floating point number.
- **Refractory period**: the time (in microseconds) during which each pixel is inactive after it has emitted an event
- **Bad pixel probability:** the probability of occurrence of hot or cold pixels can be defined in events tab. The value is given as a floating point number. The pixels are re-evaluated every time event conversion is executed. In practice, this means that each events sequence generated by the simulator looks as if it was recorded with a unique camera.
- **External noise and in-pixel noise:** modeled as time variant Gaussian noise. The noise value is drawn and applied at each potentially crossed event threshold. The value is specified as a floating point number.
- **Response time** defined as a two-pole low pass filter

- **Lens shading (vignetting):** defined as a polynomial of distance from the center point of frame F relative to frame width. This means that if we have an image with width 320 pixels, F = 1 means that a point in a frame is located 320 pixels away from the center of the image. Lens shading coefficients can be defined in events tab and are given as floating point number e.g.
- **Fixed/free running mode:** in free running mode, the event timestamps are not constrained and they can happen as often as the refractory period permits. In fixed frame-rate mode, each event has a timestamp padded to the closest allowed timestamp, and each pixel outputs only one event per frame. The frame rate is given as number of frames per second.

An example is shown in Figure 4, combining all of the above-listed characteristics to generate output events.

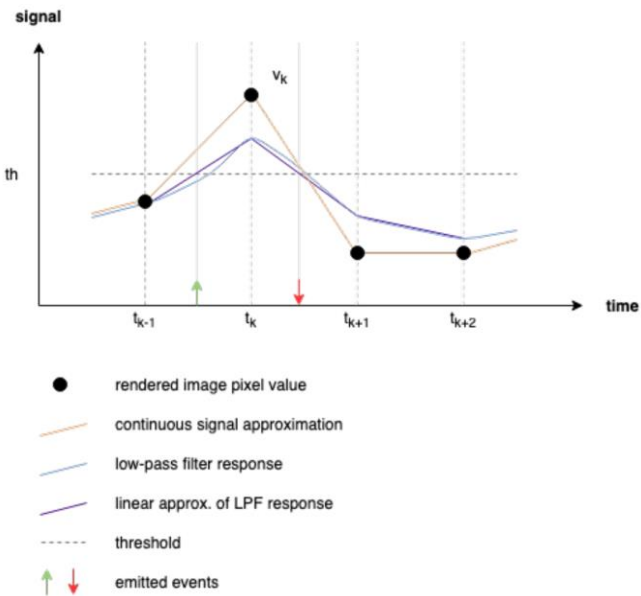

Fig. 6. Conversion example from image to events including timing estimation

The workflow pipeline of the overall simulator uses user-provided images to generate 3D textured objects, which are then placed in scenes that are then "viewed" by the virtual camera to generate events. The process is illustrated in Fig. 7.

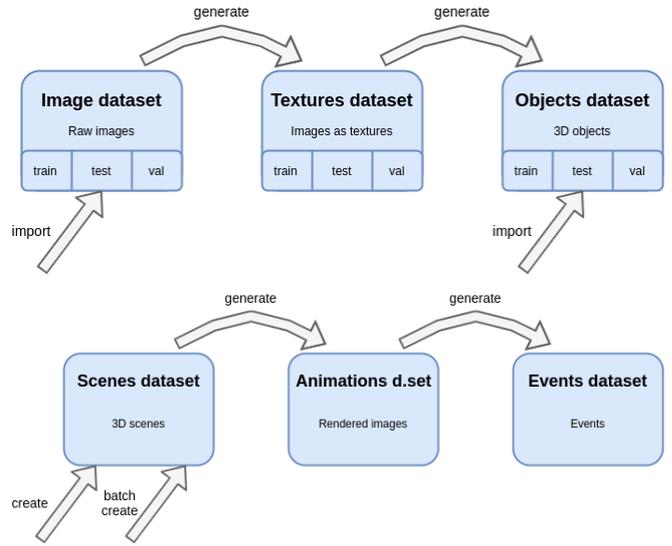

Fig. 7. Simulator workflow pipeline for (top) objects and (bottom) scenes to create events

DVS event frames are generated using the following algorithm:

1. Pre-processing

    a. Load image generated by Blender

    b. Simulate lens shading/vignetting

    c. Simulate event response time (low-pass filter) as described above

    d. Convert image to log intensity

2. Process each pixel independently of the others

    a. Check if pixel is simulated to be faulty (event always emitted, or never emitted)

    b. Check change polarity

    c. Calculate new event threshold at each pixel by combining the pixel mismatch and external noise. Pixel mismatch is modeled as a time invariant Gaussian distribution. External noise is also modeled as Gaussian noise and is applied on the event threshold every time it is recalculated.

    d. Calculate gradient of intensity change (low-pass filter response to a linear signal, see Fig. 6).

    e. Check for event threshold crossing: if it is crossed, we calculate the event timestamp using the previously calculated gradient. Using the timestamp, we check if the pixel is active or in refractory state. If the pixel is in refractory state, we only update the event threshold (no event emitted). If the pixel is active, we update the event threshold and emit an event. The process is repeated until no more thresholds are crossed.

3. Fix frame rate: In this step, all the events emitted for the most recent image frame are put in event frames. During this operation, all event timestamps are rounded to the closest event frame timestamp. It is also guaranteed that a single pixel emits only one event at the given event frame.

## B. Implementation

The simulator was built in C++ and used the Blender 3D rendering engine, as well as OpenCV and other libraries. The GUI was written using TKInter as a main library. It runs on Ubuntu Linux, although ports to other platforms are possible.

## III. IMAGE RECONSTRUCTION

### A. Network Architecture and Design

#### 1) Super Slomo

As the beginning point for our reconstruction method, we took the Super Slomo [12] network for video frame interpolation with the purpose of generating high FPS video from normal FPS (Fig. 8). Note that this network does not accept any DVS input, but we modify it for our use. The network consists of two UNets and the input is a pair of CIS images (we refer to those as the previous and next frame). Our goal is to predict a frame that occurred at any point between the two CIS frames (called the target frame).

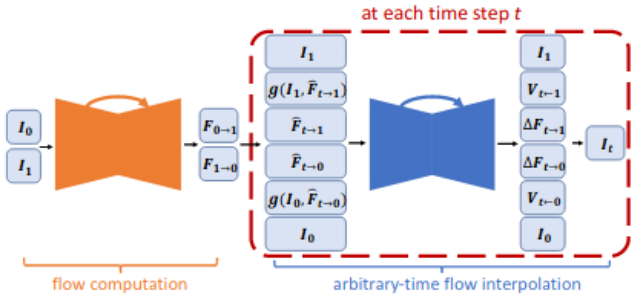

Fig. 8. Super Slomo network architecture.

In our network we follow a similar approach to the SloMo network with some modifications to accommodate the additional DVS input. As before the network consists of two UNets with the optional addition of a small convolutional recurrent neural network (RNN). The input here is a pair of CIS images and the DVS events that occurred in between them (transformed to an event tensor) as described in section 2.1.2. As a first step, the event tensor can optionally be preprocessed using a simple convolutional RNN where each bin is fed sequentially into the network to produce a tensor of the same size. This can be disabled/enabled using a hyperparameter. The procedure then follows the same steps as in SloMo with the exception that the first event bin, the last event bin, and the event bin corresponding to our target frame are used as additional input to the second UNet. The addition of the events to the second UNet allows the network to more accurately refine the flows by using information extracted from the event tensor. For example, the network might be able to predict non-linear motion between the 2 CIS frames which would not be possible without the DVS input.

The loss function used is the same as for SloMo and the same inference time optimizations are also applicable. Since the resulting image is produced by combining warped versions of the input images, at no point is the network given a chance to learn to produce HDR images given that the input is non-HDR.

#### 2) Residue Refinement Interpolation Network

The other key component of our system is based on the residue refinement interpolation network (RRIN) [13], a network for video frame interpolation that takes pairs of consecutive frames as input. It consists of 4 UNets that each perform a different function. The first UNet calculates the optical flow between the input frames. These optical flows are used to estimate the flows between the input frames and the target frame which are then refined by the second UNet. The flows are used to approximate the target frame from the input frames using backwarping which results in two versions of the target frame. The third UNet calculates a visibility mask which describes which parts of the target image was visible in previous frame and which in the next frame. This is used to linearly fuse the two approximate frames. Then the output is finally processed by the fourth UNet to produce the final frame.

The loss function used in RRIN is the Charbonnier loss which has been shown to have global smoothness properties. It is considered to be a combination of L1 and L2 loss and it is closely related to Huber loss. The loss is calculated between the predicted frame and the ground truth target frame. In contrast to the SloMo architecture, this loss is much simpler and no further restrictions regarding what the network should learn are enforced by the use of multiple loss terms. Similar to SloMo, the first network only needs to be executed once for each pair of input images during inference time.

For our network, we extend RRIN to RRINEvents, which is similarly built out of 4 UNet modules. The main difference is that this network uses event data to improve reconstruction performance. Each of the modules performs a specific task and the outputs of one module is then used as an input of another module (Fig. 9). The modules perform the following functions:

1. U1: Estimates optical flow
2. U2: Refines optical flow using event data
3. U3: Computes visibility masks (decides what part of the scene were occluded)
4. U4: Refines draft created in U3, enhances colors, and adds HDR information.

RRINEvents uses the same loss function as RRIN. Similar to RRIN, the first network only needs to be executed only once for each pair of input images during inference.

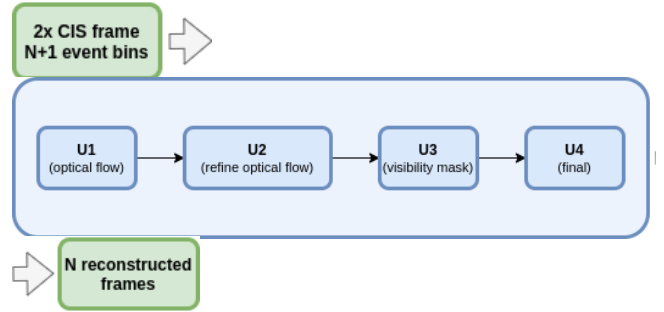

Fig. 9. RRINEvents network overview.

#### 3) Combined System

The final system, which we call RRINSloMoEvents, is a network that combines ideas from RRINEvents and SloMoEvents with the goal of reducing the number of parameters of RRINEvents without compromising performance. Just like SloMoEvents, we combine flow refinement and visibility mask calculation in the same network since those two tasks require similar inputs and could potentially benefit from sharing the same parameters. The general architecture follows that of SloMoEvents. More precisely we note the following differences:

- The event bins can be optionally preprocessed by an additional UNet which adds information taken from the CIS. The two CIS frames along with all event bins are used as input to a UNet which outputs the new event bins.
- The flow computation between the 2 CIS frames can either be done using a standard UNet which takes as input the 2 CIS frame and all event bins, or alternatively a pretrained optical flow estimation network. Since the network is pretrained, it can be frozen and no gradients need to be computed for this stage.
- There is an additional final UNet which is responsible for refining the output of the network. This also allows the network to encode HDR information to the image, something the original SloMoEvents network could not do. This step is optional. Specifically, the last UNet can be omitted (in which case the network closely resembles SloMoEvents) and HDR encoding can be done by a separate network.
- In addition to event bins, the network can accept event-only reconstructed frames as an additional input. The reconstruction can be done using either FireNet or E2VID. We reconstruct one frame per event bin in order to create a tensor of the same size as the original event tensor.

*4) HDR*

Our reconstruction networks interpolate frames independently from each other without a recurrent component to carry information across time. In addition, some of our networks are not capable of producing HDR frames since the result is always a complicating combination of the input images. For this reason, to encode HDR information we use a model originally designed for enforcing temporal consistency for independently processed video frames [14]. The goal is to run the network as a post processing step after the reconstruction networks with input the reconstructed frames and accumulated event bins, and obtain HDR temporally consistent animations as output.

The architecture is shown in Fig. 10. The network starts with 2 separate computational paths. The purpose is to make sure that event bins, which look very different from a normal image, do not directly leak into the output. The first computational path takes the reconstructed image and the previous HDR output of the model as input. The second path takes as input the event bin. After two convolutional steps, the results of the two paths are concatenated together. Skip connections are only added before the concatenation, again for the purpose of not leaking information at a late stage close to the output, and forcing the event bins to go through the full processing pipeline of the network.

In order to train this network, we input HDR frames and reconstruct frames in high dynamic range using one of our reconstruction networks. Then the resulting frames are post processed to convert them to low dynamic range conditions using random exposure settings. Specifically, for each image we randomly pick a value in the range [0, 16000] as a minimum value (any pixel value less than that will be underexposed), a slope in the range [15, 60] and noise in the range [0, 5.2]. The goal is to get back the original HDR image. The loss used is temporal consistency loss as described in the original paper, perceptual loss using VGG as the backbone network, and L1 loss.

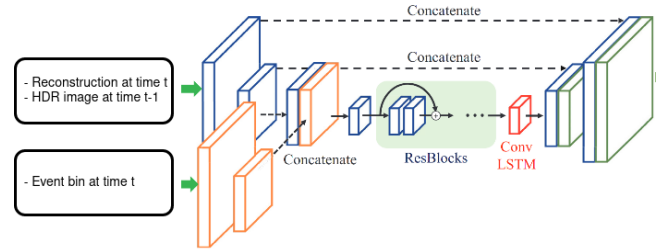

Fig. 10. Image transformation network for HDR with an LSTM component

*B. Testing: High-fps reconstruction from CIS and DVS*

Testing was performed on a Ryzen 7 CPU with 32 GB of RAM, and a RTX2080TI graphics accelerator.

The neural network used in this test use a RRIN network architecture with 17 million parameters. The main parameter of the network is fps multiplier which defines how many intermediate frames will be estimated and can be set to any (reasonable) value. We used a multiplier of 32 on a 30 fps CIS input, which resulted in 960fps output videos.

TABLE I. TRAINING DATA SUMMARY

| Source | Simulator | Cameras |
| --- | --- | --- |
| # scenes | 180 | 960 |
| CIS fps | 30 | 30 |
| Interpolated fps | 960 | 960 |
| # samples | 34560 | 184704 |
| DVS/CIS disparity | <= 15 mm | 0 mm |

*1) Source data*

The simulated data was created from scratch in the Event Simulator using random images as textures of objects. The rendered scenes contain random shapes moving in front of a camera. Each of the scene has a randomly offset DVS sensor by a maximum distance of 15mm from the CIS sensor.

The rest of the scenes were recorded using various Full-HD cameras (GoPro, smartphones) and stored in a format consistent with the event simulator. Post processing and converting images to DVS event data was performed in the simulator.

The original videos contain slowly moving objects and were recorded at 240fps, and then accelerated to 960fps. This means that a car driving 10km/h will look as if it was driving 40km/h in the accelerated video.

Due to post-processing limitations, we were not able to introduce disparity between the CIS and DVS sensor in the Full-HD camera recordings.

*2) Post processing*

All the animations were post-processed in the simulator using the following parameters:

- CIS animation

- Minimum detected illuminance 4096
- Slope 55
- Noise 5.2
- DVS animation
  - Event threshold 0.15
  - Pixel mismatch 0.015
  - Hot/cold pixel probability 0.0
  - External noise 0.035
  - Refractory period 100

*3) Reconstruction results*

The reconstruction required 302 s per frame of reconstructed video. This included the time to read data from storage, reconstruct the images, and store the results.

Two test sets were used: a 3D simulated environment and a set from recorded video that was then converted into synthetic events at simulated high speed. The overall statistics are shown in TABLE II. Sample output stills are show in Fig. 11 and Fig. 12, and a graphical sample of reconstruction errors is shown in Fig. 13.

TABLE II.   RECONSTRUCTION IMAGE QUALITY

| SSIM | Simulated | Recorded |
| --- | --- | --- |
| Mean SSIM | 0.955 | 0.937 |
| Min SSIM | 0.941 | 0.888 |
| Min. patch SSIM (99 x 99 pixels) | 0.478 | 0.500 |
| Std. dev. SSIM | 0.006 | 0.014 |

| PSNR | Simulated | Recorded |
| --- | --- | --- |
| Mean PSNR | 38.122 | 34.865 |
| Min PSNR | 34.964 | 28.441 |
| Min. patch PSNR (99 x 99 pixels) | 22.493 | 15.404 |
| Std. dev. PSNR | 1.688 | 2.245 |

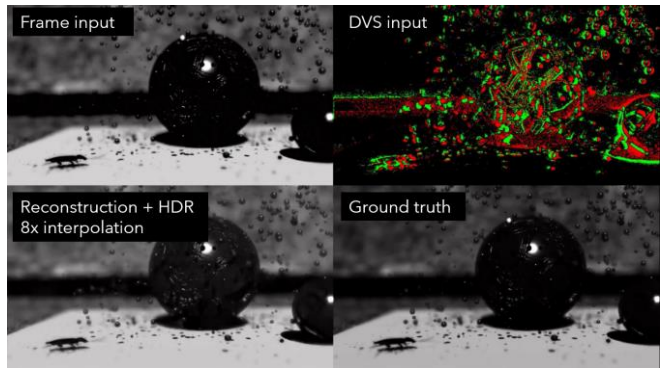

Fig. 11. Still frame from video of reconstruction demonstration using synthetic 3D environment. (top left) CIS frame input; (top right) DVS input; (bottom left) reconstructed scene; (bottom right) original ground truth.

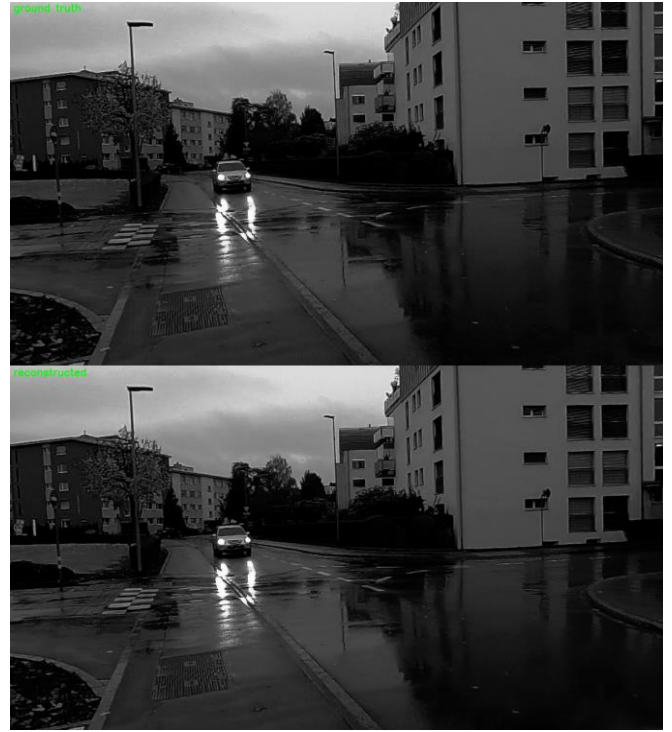

Fig. 12. Still frame from video of reconstruction from real video: (top) original; (bottom) reconstructed.

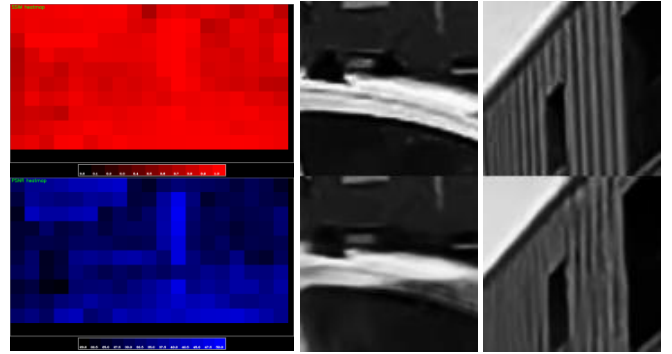

Fig. 13. Left: Sample reconstruction quality heatmap as measured by SSIM (left top) and PSNR (left bottom). Middle: Example worst-case PSNR reconstruction image patch; original (top) vs reconstructed (bottom). Right: Example worst-case SSIM reconstruction image patch; original (top) vs reconstructed (bottom).

*4) Error source analysis*

It is instructive to look at details of instances where the reconstruction yielded low quality, and the causes for these effects.

*a) Sensor disparity*

The most surprising difficulty was the unexpectedly high impact of the disparity between the DVS and the frame-based sensors. An imperfect geometrical location of the sensors and imperfect time-synchronization between them causes difficulties in reconstructing HDR videos with high motion speed. This means that the optical flow cannot be correctly determined when the disparity between the sensors is too high. The result is warped frames and degraded quality (see Fig. 14 for an example).

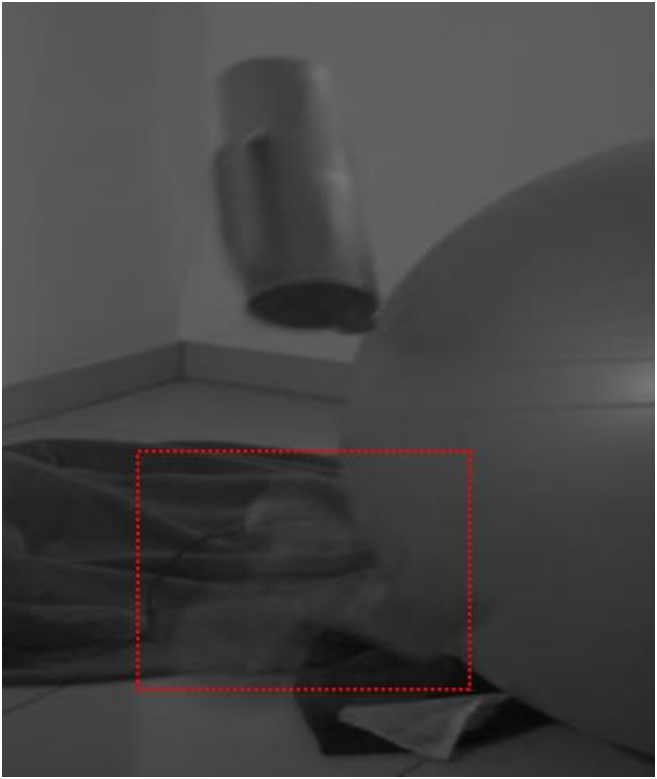

Fig. 14. Example of disparity, fast motion and spinning of an object causing strong artifacts. The object on the top, which moves slower, was reconstructed correctly, but the object at the bottom suffers from ghosting effects.

The sensor disparity also affects the quality of HDR improvements. Dynamic range can be improved using events located near the CIS pixel. However, error in pixel locations due to disparity can led to strong edge artifacts in some scenes. Fig. 15 illustrates one example of such a case.

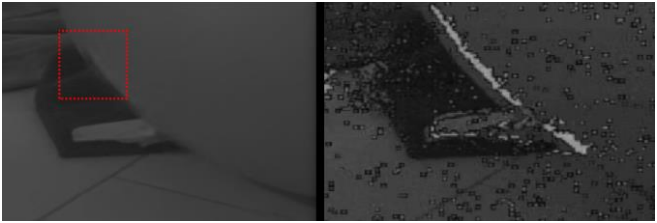

Fig. 15. Example of HDR enhancement near the edge of a ball. Due to disparity errors, the pixel values were changed in the wrong location leading to double-edge errors.

#### b) Blurry frame input

Our neural network architecture depends on warping the input images based on optical flow. The DVS is less prone to blur than frame sensors, which means that a combination of the two data streams could potentially remove the motion blur from the reconstructed videos. There has been some work on using events to reduce motion blur [15] which was not integrated into our model.

#### c) Very fast motion

One of the base components of our reconstruction network is an optical flow estimator. This estimator searches for correspondences between two consecutive CIS input images. If this component fails due to excessive motion, the reconstructed images have strong warping artifacts. There exists a limit of maximum speed of motion which can be reconstructed. This limit depends on the event data quality, texture and other parameters. The artifacts caused by fast motion are also present in simulated data. Fig. 16 shows a mild example of such artifacts, while Fig. 17 shows a more extreme example.

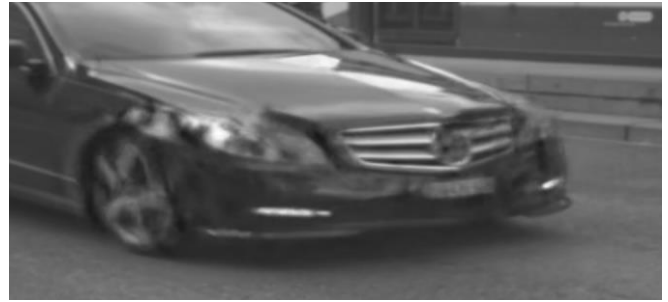

Fig. 16. Example reconstruction artifacts in a car driving by with a high speed and low image contrast

The fast motion artifacts occur because the network cannot reconstruct motion bigger than its receptive field. This means, that the algorithm can only reconstruct objects which move less than X pixels between two frames. Objects which move faster will be reconstructed with strong artifacts (see Fig. 17 for an example). The reason why the network receptive field is limited is the network kernel sizes. The bigger the kernels, the bigger motion the network can understand, at the cost of required computational power. The current model allows high-quality reconstruction at up to ~100 pixels per frame (3000px/s).

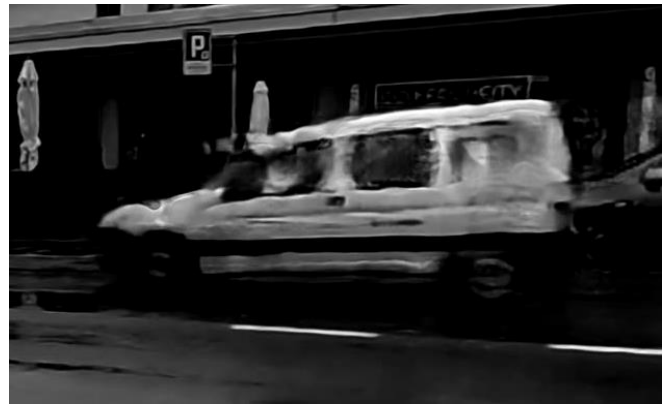

Fig. 17. Example reconstruction artifact from a high-speed car moving at an equivalent speed of 7500 pixels/s.

#### 5) Comparison with Time Lens

At the moment, there are very few similar solutions and publicly available datasets available for direct evaluation. Our neural network combining images and events is not directly comparable with frame-only interpolation networks (e.g. Super SloMo [12]), as our solution has more input event information. Currently, the only other known work we can compare with is Time Lens [16], which also provides a dataset (High Speed Event-RGB). There are, however, some differences between the two approaches worth mentioning:

1. Disparity learning: Time Lens pre-warps the images to be aligned with the events, so its network did not have to learn any disparity.

2. Resolution matching: Time Lens assumes a matched DVS and frame resolution (1:1). Our ratio was not fixed (in this case it was 1:9), meaning that our model had to learn its general neighborhood context.

We compared Time Lens with our solution using the HS-ERGB dataset by increasing the frame rate of the reconstructed video by factor of 8. This means that we only used every 8th CIS frame as the input to the network (skip=7, insert=7). The evaluation was conducted using two methods – firstly the one provided by Time Lens, and secondly by our own evaluation framework which produces more detailed visualizations. As seen in TABLE III. , our network achieves the same PSNR results as Time Lens, and slightly better SSIM results.

TABLE III. Image reconstruction evlaution (all 15 recordings of HS-ERGB dataset)

| Method | SSIM | PSNR |
| --- | --- | --- |
| Time Lens | 0.869±0.123 | 32.207±7.18 |
| Ours | **0.877±0.109** | **32.205±7.04** |

In order to understand the visual quality of the metrics, we analyzed the videos as a function of time. We can see that up to a certain speed of motion (measured in pixels per second) we can completely reconstruct the frame with very low quality loss. On the other hand, when the motion is too fast our score metrics begin to drop.

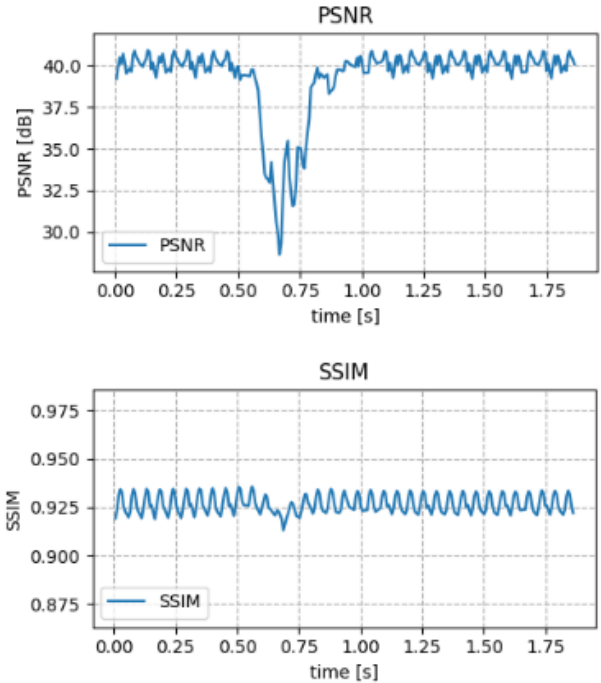

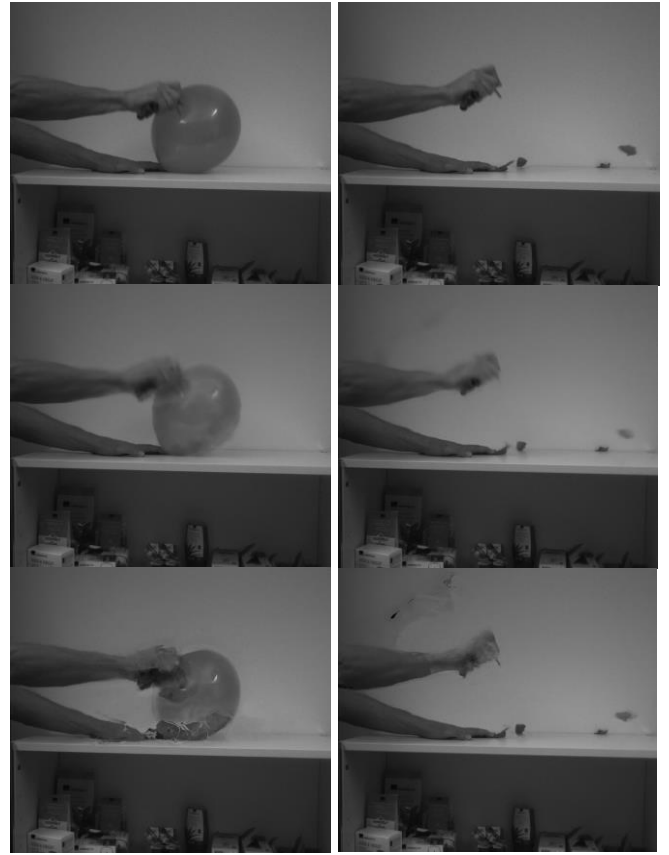

Fig. 19. Artifacts produced by our network and Time lens in a balloon popping scene. Left column: balloon just at the point of popping; right column: arm being retracted just after balloon is popped. Top row: ground truth; middle row: our reconstruction; bottom row; Time Lens reconstruction. Left: Our reconstruction shows blurring artifacts of the balloon and hand just before popping, while Time Lens shows sharper artifacts on the balloon and hand. Right: Our reconstruction shows a faint blur artifact of the arm in the top-left corner before it arrives in that location, while Time Lens shows sharp artifacts of the arm in the same location.

Fig. 18. Quality metrics as a function of time for balloon popping video. The period of quality reduction corresponds to very fast motion, i.e. the balloon popping. The quality then recovers to a higher level when there is no longer fast motion in the scene.

Our network and the Time Lens reconstructions generally showed the same locations of degraded quality, but with different emphases. Our network generally tended towards blurred artifacts, while Time Lens exhibited sharper artifacts. (Fig. 19). We suggest that these differences may be due to different weightings in forward/backward motion prediction.

IV. Discussion, Conculsions, and Further Work

The motivation for us to include many noise parameters in our event+frame simulator was to provide raw output that closely matched real sensors, from a statistical point of view. Having access to this level of realism was important for our particular application of image interpolation. For other applications where human interpretation of the output is not critical (e.g. visual odometry), simpler models are likely to be sufficient in many cases.

Our network training method combined relatively simple 3D environments with high-resolution photorealistic textures for training. Despite the missing moving parallax information using this method, we achieved good reconstruction results.

This work has shown that combining DVS with CIS can create high-quality image reconstruction, provided that the motion in the image remains within certain bounds so that the neural network receptive fields are able to perform inference effectively. Having access to a full-featured simulator with realistic noise and image artifact models was a crucial part of the process of developing and testing the network.

This work set many goals to be achieved in parallel by the same network:

1. Reconstruct high fps video in end-to-end manner, with minimal assumptions about sensor resolution and disparity

2. Improve dynamic range

3. Deblur the frames

4. Be resistant to flickering

The interference between these goals is significant and it is highly challenging to achieve all of them at once, because these tasks compete for the computational capacity of the network. For example, training the network with different dynamic range of input and output reduces the quality of the optical flow reconstruction. Future approaches may focus on each of these elements separately, enabling a combined solution with different weightings depending on the image context. Furthermore, ideas could be drawn from the multi-task learning literature for deep learning, to correctly identify which subset of tasks can benefit from joint training and which ones can hurt model performance.

The reconstructed images were not gamma-corrected. Reconstructing gamma-corrected images would mean that the CIS and interpolated animations produced by the simulator would be gamma-corrected, but DVS animation would be output raw. This would make the results much more pleasing to the human eye and easier to evaluate subjectively.

ACKNOWLEDGMENT*S*

We would like to thank Samsung Electronics VP Jesuk Lee for his support, and Dr. Junseok Kim and Dr. Yunjae Suh for providing DVS characteristics.